%% file: samplepaper.tex
\documentclass{llncs}
\usepackage{url}
\usepackage{comment}
\usepackage{amsfonts}
\usepackage{amsmath}
\usepackage{amssymb}
\usepackage[linesnumbered,algoruled,boxed,vlined, noend]{algorithm2e}
\usepackage{subcaption}
\usepackage{graphicx}
\usepackage{optidef}
\usepackage{ulem}
\usepackage{makecell}
\usepackage[table]{xcolor}

\usepackage{wrapfig}
\usepackage{overpic}
\usepackage[cmtip,arrow]{xy}
\usepackage{pb-diagram,pb-xy}
\usepackage{tikz}

\input{defs}

\newcommand{\aravind}[1]{ {\color{red}#1} }

\DeclareTextFontCommand{\fancyEmph}{\bfseries\em}

\input{correctm}

\showcomments

\begin{document}
\mainmatter              
\title{Morse Graphs: Topological Tools for Analyzing the Global Dynamics of Robot Controllers}
\titlerunning{\aravind{Running title here}}  
%
\author{Ewerton R. Vieira\inst{3,5}, Edgar Granados\inst{1}, Aravind Sivaramakrishnan\inst{1},\\ Marcio Gameiro\inst{2,4}, Konstantin Mischaikow\inst{2}, Kostas E. Bekris\inst{1} \footnote{The work is supported in part by an NSF HDR TRIPODS award 1934924. MG and KM were partially supported by the NSF under awards DMS-1839294, DARPA contract HR0011-16-2-0033, and NIH award R01 GM126555. MG was also partially supported by CNPq grant 309073/2019-7.}}
\authorrunning{\aravind{Running author list here}} 
%
\tocauthor{\aravind{TOC author list here}}
%
\institute{Dept. of Computer Science, Rutgers University, NJ, USA\\
\and
Dept. of Mathematics, Rutgers University, NJ, USA\\
\and
DIMACS, 
 Rutgers University, NJ, USA \\
\and
ICMC, Universidade de S\~{a}o Paulo, S\~{a}o Carlos, S\~{a}o Paulo, Brazil
\and
IME, Universidade Federal de Goi\'{a}s, Goi\^{a}nia, GO, Brazil.\\
\email{\{er691,eg585,as2578,gameiro,mischaik,kostas.bekris\}@rutgers.edu}
}

\maketitle              

\vspace{-.2in}
\begin{abstract}
Understanding the global dynamics of a robot controller, such as identifying attractors and their regions of attraction (RoA), is important for safe deployment and synthesizing more effective hybrid controllers.  This paper proposes a topological framework to analyze the global dynamics of robot controllers, even data-driven ones, in an effective and explainable way. It builds a combinatorial representation representing the underlying system’s state space and non-linear dynamics, which is summarized in a directed acyclic graph, the \textit{Morse graph}. The approach only probes the dynamics locally by forward propagating short trajectories over a state-space discretization, which needs to be a Lipschitz-continuous function.  The framework is evaluated given either numerical or data-driven controllers for classical robotic benchmarks. It is compared against established analytical and recent machine learning alternatives for estimating the RoAs of such controllers. It is shown to outperform them in accuracy and efficiency. It also provides deeper insights as it describes the global dynamics up to the discretization’s resolution.  This allows to use the Morse graph to identify how to synthesize controllers to form improved hybrid solutions or how to identify the physical limitations of a robotic system.
\keywords{Topology, Robot Control, Robot Dynamics.}
\end{abstract}
\vspace{-.3in}

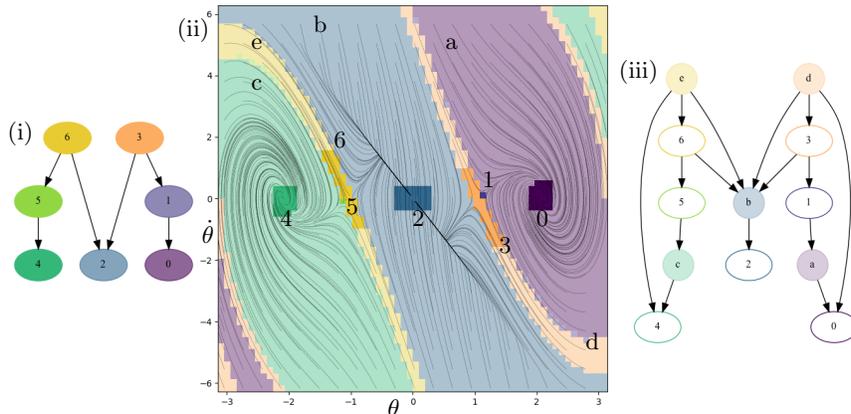
\begin{figure}[t]
\centering
\hspace{-0.7cm}\begin{overpic}[width=0.45\columnwidth]{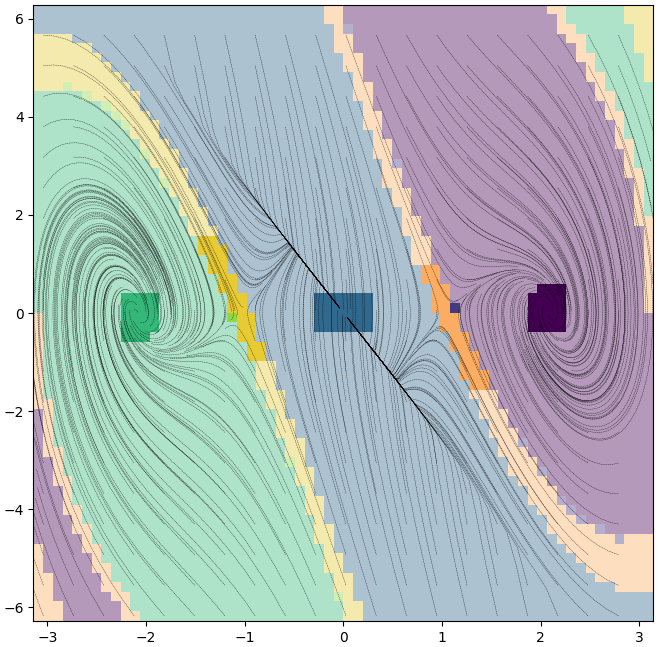}
\put(-5, 90){(ii)}
\put(13, 87){e}
\put(28, 91){b}
\put(13,77){c}
\put(60,87){a}
\put(94,14){d}
\put(20,44){\small $4$}
\put(36,47){\small $5$}
\put(33,63){\small $6$}
\put(52,44){\small $2$}
\put(73,37.5){\small $3$}
\put(69,53.5){\small $1$}
\put(82,44){\small $0$}


\put(46,-2){\small $\theta$}
\put(1,40){\small $\dot\theta$}

\put(-45,30){\includegraphics[width=0.2\columnwidth]{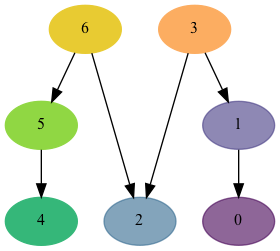}}
\put(-46,65){(i)}

\put(105,15){\includegraphics[width=0.25\columnwidth]{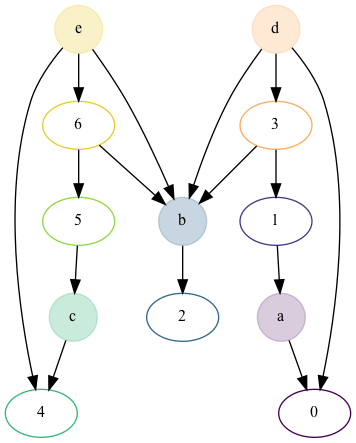}}
\put(102,80){(iii)}

\end{overpic}


\vspace{-.1in}
\caption{\small Results of the proposed combinatorial analysis for a pendulum operating under an LQR controller so as to reach the (0,0) state in the phase space. (i) The method outputs the Morse graph $\sMG(\cF)$ (left) to identify recurrent dynamics. (ii) The numbered Morse cells $\sM$ represent recurrent sets associated with nodes of $\sMG(\cF)$. The RoAs of the regions in $\sM$ are indicated by the corresponding light colors and letters. (iii) The graph on the right is used to identify the RoAs for the corresponding recurrent sets.}
\vspace{-.3in}
\label{fig:pendulum12}
\end{figure}

\section{Introduction}
\label{sec:introduction}
\vspace{-.05in}


Estimating Regions of Attraction (RoA) of a dynamical system is needed in robotics for understanding the conditions under which a controller can be safely applied to solve a task. It is also needed for composing controllers and forming hybrid solutions that work from a wider swath of the underlying state space.  Estimating such RoAs, however, is challenging. 

Computing a Lyapunov function (LF) can provide an RoA but obtaining an analytical expression for an LF is difficult for general non-linear systems. This motivated numerical solutions, which often still require access to the system's differential equation for computing an LF. Recent advances in data-driven control provide effective learned controllers \cite{Haarnoja2018SoftAO,gillen2020combining}, which do not have analytical expressions. Machine learning methods have also been proposed to learn RoAs \cite{chen2021learning_nonlinear,berkenkamp2016safe} without access to the control law's expression as well as for composing controllers \cite{perkins2002lyapunov,7330913}. These methods, however, tend to be sensitive to parameters, are computationally demanding in time and memory while lacking guarantees. Therefore, there is a need for identifying RoAs, especially of data-driven systems, in a robust, mathematically rigorous, and computationally efficient way.

{\bf Contributions:} This work sidesteps the estimation of an LF.  It builds on top of recent progress in combinatorial dynamics and order theory \cite{kalies:mischaikow:vandervorst:14,kalies:mischaikow:vandervorst:15,kalies:mischaikow:vandervorst:21} to propose a combinatorial analysis of the global dynamics of black-box robot controllers and describe attractors and their corresponding RoAs. The approach is based on a finite combinatorial representation of the state space and its nonlinear dynamics. It only requires access to a discrete time representation of the dynamics and can handle analytical, data-driven and hybrid controllers. 

For instance, consider the phase space of a pendulum as in Fig. \ref{fig:pendulum12} given an LQR controller for driving the system to the $(0,0)$ state. The proposed framework generates the combinatorial representation on the right, where nodes correspond to regions in a state space decomposition. This information is then summarized in more compact, annotated, acyclic directed graphs called Morse graphs, shown on the left. The nodes of Morse graphs can contain attractors of interest. The associated RoAs can also be inferred automatically from the combinatorial representation. The resulting information is finite and graphical in nature, thus, it can be easily queried and understood by a person. The accompanying evaluation shows that the proposed tools are relatively computationally efficient, provide a more global, explainable understanding of the dynamics, often achieve higher accuracy and provide stronger guarantees than alternatives. They also allow the composition of hybrid controllers with wider RoAs.

{\bf Related Work:} Multiple numerical methods exist for estimating RoAs given direct access to the system's expression \cite{giesl2015review}. 
For instance, maximal Lyapunov functions (LFs) \cite{vannelli1985maximal} incrementally compute the RoA. 
Constructing an ellipsoidal RoA approximation reduces to a linear matrix inequalities (LMIs) problem \cite{pesterev2017attraction,pesterev2019estimation}, which has been applied to wheeled robots \cite{rapoport2008estimation} and NASA's generic transport model \cite{pandita2009reachability}.
There are also convex formulations that rely on LMI relaxations to solve a convex linear program and approximate the RoA of systems with polynomial dynamics and semi-algebraic inputs \cite{henrion2013convex}. An LF can be restricted to be a sum-of-squares (SoS) polynomial constructed via semi-definite programming \cite{parrilo2000structured}. SoS methods can build randomized stabilized trees with LQR feedback \cite{tedrake2010lqr}, pre-compute funnel libraries \cite{majumdar2017funnel} and acquire certificates of stability of rigid bodies with impacts and friction \cite{posa2013lyapunov}. State space samples satisfying a Lyapunov-type inequality can construct neighborhoods where the candidate LF is certified \cite{bobiti2016sampling}. In contrast, the proposed approach avoids computing an LF and does not require access to the analytical expression of the underlying control law. 

Reachability analysis \cite{bansal2017hamilton} computes the backward reachable tube of a dynamical system returning the maximal RoA without imposing shape restrictions. Tasks include RoA computation of dynamical walkers \cite{choi2022computation} and, combined with machine learning, maintain the system's safety over a given horizon \cite{gillulay2011guaranteed}. Barrier functions ensure safety of unknown dynamical systems, and can be learned with Gaussian Processes (GP) to obtain safe policies \cite{akametalu2014reachability}. Barrier certificates (BCs) can identify areas needing exploration to expand the safe set~\cite{wang2018safe}.  

Machine learning can be used to compute both LFs and BCs. One approach is to alternate between a learner and a verifier to search within a LFs set \cite{chen2021learning_hybrid}. An alternative approximates the dynamics map as a piecewise linear neural network using a counterexample-guided method as a verifier to synthesize a LF \cite{chen2021learning_nonlinear}. LFs can also be constructed via stable data-driven Koopman operators \cite{mamakoukas2020learning}. LFs and BCs can be obtained by training a neural network and using an SMT solver as a verifier \cite{abate2021fossil}. LFs for piecewise linear dynamical systems can be synthesized as the outputs of neural networks with leaky ReLU activations \cite{dai2020counter}. Given a system's initial safe set, a neural network LF is trained to adapt to the RoA's shape \cite{richards2018lyapunov}. GPs can be used to obtain a Lyapunov-like function \cite{lederer2019local}. Finally, LFs can be synthesized while learning a controller to prove the controller's stability and generate counter-examples to improve the controller \cite{dai2021lyapunov}. This paper compares performance against a state-of-the-art ML approach that computes an LF \cite{richards2018lyapunov}.

This work builds on top of topological tools. Topology has been used for various problems in robotics, such as deformable manipulation \cite{bhattacharya2015topological,antonova2021sequential}, robot perception \cite{ge2021enhancing}, multi-robot problems \cite{varava2017herding}, determining homotopy-inequivalent trajectories \cite{pokorny2016high} and to extract higher-order dynamics for motion prediction \cite{carvalho2019long}. Morse theory has been used to incrementally build local minima trees for multi-robot planning problems \cite{orthey2020visualizing}, and to find paths to cover 2D and 3D spaces \cite{acar2002morse}. To the best of the authors' knowledge, this is the first application of recent advancements in topology to summarize the global dynamics of robot controllers. 
\vspace{-.15in}

\section{Problem Setup}
\label{sec:setup}
\vspace{-.05in}

This work aims to systematically analyze the global dynamics of robot controllers based on combinatorial dynamics and order theory \cite{kalies:mischaikow:vandervorst:14,kalies:mischaikow:vandervorst:15,kalies:mischaikow:vandervorst:21}. The prior theory is very general and applies to any continuous dynamical system defined over a locally compact metric space. The material is adopted and applied to the restricted setting of robot control problems. In particular, consider a nonlinear continuous-time system:
    \vspace{-.15in}
\begin{equation}\label{eq:dyn}
    \dot{x} = f(x,u),
    \vspace{-.05in}
\end{equation}
where $x(t)\in X$ is the state at time $t$ in a domain $X\subseteq \mathbb{R}^n$, $u: X \rightarrow \mathbb{U} \subseteq \mathbb{R}^m$ is a Lipschitiz continuous control as defined by a control policy $u(x)$, and $f:X\times \mathbb{U} \rightarrow \mathbb{R}^n$ is a Lipschitiz continuous function, where $\mathbb{U}$ is an open set in $\mathbb{R}^m$. The dynamical system consists of the model $f(\cdot)$, which can be accessed but it is not necessarily known analytically, and a control policy $u=u(x)$ that can be either analytical or learned from data. 

For a given time $\tau >0$, let $\phi_\tau: X \rightarrow X$ denote the function derived from solving  Eq. \eqref{eq:dyn} forward in time for duration $\tau$ from everywhere. Given that $f$ and $u$ are Lipschitz continuous, $\phi_\tau$ is also Lipschitz continuous. Denote the global Lipschitz constant of $\phi_\tau$ by $L_\tau$.  Observe that a RoA for Eq. \eqref{eq:dyn} is a RoA under $\phi_\tau$. Therefore and w.l.o.g, the rest of this work focuses on the dynamics of $\phi_\tau$, which is not assumed, however, to be computable and available.




The objective is to identify a combinatorial approach, which can capture meaningful aspects of the dynamics of interest according to $\phi_\tau\colon X\to X$, which are continuous in nature. In this context, a subset of the state space $N\subset X$ is an \emph{attracting block} for $\phi_\tau$, if $\phi_\tau(N)\subset \mathrm{int}(N)$, where $\mathrm{int}$ denotes topological interior. This means that the system of Eq. (1) will not escape the subset $N$ once it has entered it.
Denote the set of attracting blocks of $\phi_\tau$ by $\sABlock(\phi_\tau)$.
Given $N\in \sABlock(\phi_\tau)$, its \emph{omega limit set} is an invariant set for $\phi_\tau$ defined as: \vspace{-.1in}
\[
\omega(N):= \bigcap_{n \in \mathbb{Z}^+} \mathrm{cl}\left(\bigcup_{k=n}^\infty \phi_\tau^k(N)\right)
\vspace{-.125in}
\]
where $\phi_\tau^k$ is the composition $\phi_\tau\circ \cdots \circ\phi_\tau$ ($k$ times) and $\mathrm{cl}$ is topological closure. \emph{The attracting block $N$ is a RoA for $\omega(N)$}. In general $\sABlock(\phi_\tau)$ is huge, containing uncountably many elements, and is too large to work with. 
Thus, the {\bf problem} is to systematically identify a minimal finite subset of $\sABlock(\phi_\tau)$ that both represents as tightly as possible the attractors and captures the maximal RoAs of these attractors.

{\bf Running Example:} For exposition purposes, the following discussion will use the second-order pendulum as an example to explain the corresponding definitions and the proposed method (Fig~\ref{fig:pendulum12}). The pendulum is modeled by the differential equation $m\ell^2 \Ddot{\theta} = mG \ell \sin{\theta} - \beta\theta + u$, with state $x := (\theta, \dot\theta)$, where $\theta$ is the angle from the upright equilibrium $\mathbf{\theta}_o = 0$, $u$ is the input torque, $m$ is the pendulum mass, $G$ is the gravitational acceleration, $\ell$ is the pole length, and $\beta$ is the friction coefficient. The control $u$ in the running example is computed by the LQR approach described in Section \ref{sec:results}. In Fig.~\ref{fig:pendulum12}, we use the time-$1$ map $\phi_1$ of the flow of the pendulum under the LQR controller. The proposed method is not limited to this or similar low-dimensional systems/controllers. 
\vspace{-.15in}

\section{Proposed Framework and Method}
\label{sec:FM}
\vspace{-.05in}

{\bf Overview:} The method first approximates $\phi_\tau$ by decomposing the state space $X$ into regions $\xi$. For multiple initial states within each $\xi$ the system is propagated forward for time $\tau$  to identify regions reachable from $\xi$. Given the reachability information and the Lipschitz continuity of $\phi_\tau$, a directed multi-valued graph representation $\cF$ stores each region $\xi$ as a vertex and edges point from $\xi$ to all regions in an outer (conservative) approximation of its true reachable set.

The method then computes the strongly connected components (SCC) of $\cF$. An SCC is a maximal set of vertices of $\cF$ such that every pair of vertices in the SCC are reachable from each other. The non-trivial SCCs of $\cF$, i.e., those with at least one edge, are called \emph{recurrent sets}, and capture the recurrent dynamics of $\phi_\tau$. Every region $\xi$ not in a recurrent set exhibits non-recurrent behavior.  The same algorithm that computes SCCs also provides a topological sort of the vertices in $\cF$, which allows to define reachability relationships between recurrent sets and non-recurrent regions.  This gives rise to a condensation graph $\sCG(\cF)$, where all SCCs are condensed to a single vertex and edges reflect reachability according to the topological sort. Typically, $\sCG(\cF)$ is roughly the same size as $\cF$ and cumbersome to maintain. The implementation avoids explicitly storing either graph.  The method succinctly captures the recurrent and non-recurrent dynamics in the Morse graph $\sMG(\cF)$, whose vertices are the recurrent sets of $\cF$ and whose edges reflect reachability according to the topological sort. Overall, the proposed method can be divided into the four steps described below:  \vspace{-.1in}
\begin{itemize}
    \item {\bf Step 1.} State space decomposition and generation of input to represent $\phi_\tau$;
    \item {\bf Step 2.} Construction of the combinatorial representation $\cF$ of the dynamics given an outer approximation of $\phi_\tau$;
    \item {\bf Step 3.} Computation of Condensation Graph $\sCG(\cF)$ and Morse Graph $\sMG(\cF)$ via identification of recurrent sets/SCCs of $\cF$ and topological sort;
    \item {\bf Step 4.} Derivation of RoAs for the recurrent sets from $\sCG(\cF)$;
\end{itemize}

\vspace{-.05in}
\noindent {\bf Step 1 - State Space Decomposition and Generation of Input Data:} This paper considers the control system of Eq. \eqref{eq:dyn} restricted to a state space given by an orthotope $X = \prod_{i=1}^n [a_i, b_i]$, allowing for the possibility of periodic boundary conditions. This allows for torus-like spaces, such as for the running example of the 2$^{nd}$-order pendulum. For simplicity, the accompanying implementation is using a uniform discretization of the state space based on $2^{k_i}$ subdivisions in the $i$-th component resulting in a decomposition of the state space into $\prod_{i=1}^n 2^{k_i}$ cubes of dimension $n$. The term $\cX$ denotes the collection of these cubes. 

The method then generates the set of values of $\phi_\tau$ at the corner points of cubes in $\cX$. More precisely, let $V(\cX)$ denote the set of all corner points of cubes in $\cX$. The method computes the set of ordered pairs $\Phi_\tau(\cX):=\{(v,\phi_\tau(v)) \mid v \in V(\cX)\}$, by forward propagating the dynamics for time $\tau$
from all $V(\cX)$. In this way, this work does not assume exact, analytical knowledge of $\phi_\tau$, but rather exploits its existence. It only requires the ability to generate the set $\Phi_\tau(\cX)$.



\begin{wrapfigure}{r}{0.6\textwidth}
\vspace{-.3in}
\includegraphics[width=0.295\columnwidth]{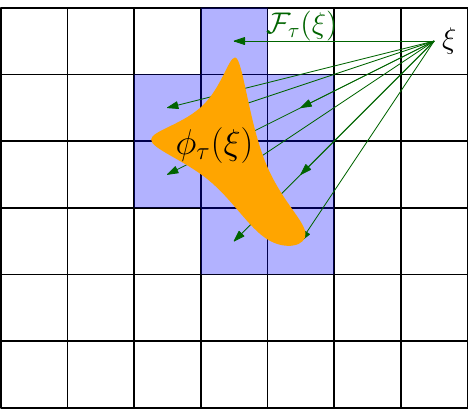}
\includegraphics[width=0.295\columnwidth]{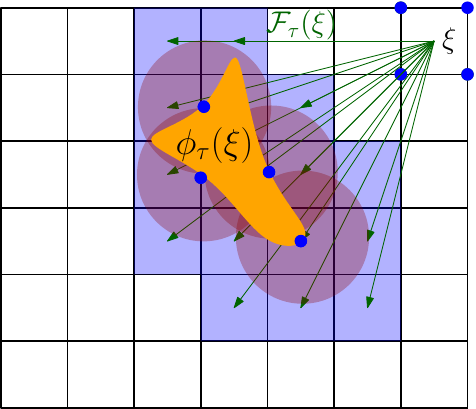}
\vspace{-0.25in}
\caption{\small The set $\phi_{\tau}(\xi)$ denotes the reachable states from all states in the cell $\xi$ after time $\tau$. Multivalued maps: (left) minimal/ideal outer approximation $\cF_{min}(\xi)$ and (right) outer approximation obtained by using a Lipschitz constant.}
\vspace{-.3in}
\label{fig:outer_approx}
\end{wrapfigure}

\noindent {\bf Step 2 - Combinatorial Representation $\cF$ via outer approximation:} 
The dynamics of the continuous $\phi_\tau$ are approximated by a \emph{combinatorial multivalued map} $\cF \colon \cX\mvmap \cX$, where vertices are $n$-cubes $\xi \in \cX$. The map $\cF$ contains directed edges $\xi \to \xi', \forall\ \xi' \in \Phi_\tau(\xi)$.
The set of cubes identified by $\cF(\xi)$ are meant to capture the possible states of $\phi_\tau(\xi)$. To obtain mathematically rigorous results about the dynamics of $\phi_\tau$, it is sufficient for $\cF$ to be an \emph{outer approximation} of $\phi_\tau$, i.e., for 
\vspace{-.125in}
\begin{equation}
    \label{eq:OuterApproximation}
    \phi_\tau(\xi) \subset \mathrm{int}\left(\cF(\xi)\right)\quad \text{for all $\xi\in\cX$}
\vspace{-.025in}
\end{equation}
where $\mathrm{int}$ denotes topological interior. 
The left side of Eq. \eqref{eq:OuterApproximation} indicates the set of states that can be achieved in time $\tau$ according to \eqref{eq:dyn}, which is unknown exactly. On the right side, $\cF(\xi)$ is a list of vertices that can be identified with a set of $n$-cubes in $X$. 
The inclusion relation and $\mathrm{int}$ indicate the constraint that a large enough collection of cubes is chosen in $\cF(\xi)$ to enclose $\phi_\tau(\xi)$ and at least an arbitrarily small overestimation is needed. The minimal outer approximation of $\phi_\tau$ \cite{kalies:mischaikow:vandervorst:05} is:
\vspace{-.125in}
\[
\cF_{min}(\xi) := \{ \xi' \in \cX  \mid  \xi'\cap \phi_\tau(\xi) \neq \emptyset \},
\vspace{-.025in}
\]
that is, $\cF_{min}(\xi)$ indicates the minimal set of cubes that contain the set of all states that can be reached in time $\tau$ starting in $\xi$. See Fig. \ref{fig:outer_approx}(left) for an illustration.  Even with complete knowledge of $\phi_\tau$, computation of $\cF_{min}$ is typically prohibitively expensive. Nevertheless, from a mathematical perspective it suffices to work with any $\cF\colon \cX\mvmap \cX$ that satisfies $\cF_{min}(\xi) \subset \cF(\xi)$ for all $\xi\in\cX$. In general, the objective is to achieve a tight outer approximation, as the larger the size of the images of $\cF$, the less the information about the dynamics of $\phi_\tau$. 

The assumption for an outer approximation is that $\cF(\xi)\neq\emptyset$ for all $\xi\in\cX$. In practice it does not need to hold. Determining $\cF$ represents the major computational bottleneck as it involves numerical simulations of Eq. \eqref{eq:dyn} or obtaining real-world experiments with the robotic system. The flexibility in the definition of an outer approximation provides flexibility in its construction. This work computes $\cF$ as follows: Recall that $\phi_\tau$ is Lipschitz with constant $L = L_{\tau}$ ($L$ depends on $\tau$). Given $x\in X$,  let $\overline{B(x,\delta)}  = \{x'\in X\mid \|x-x'\| \leq \delta \}$ denote the $\delta$-closed ball at state $x$.
Define the \emph{diameter of $\xi \in \cX$} by $d(\xi) := \max_{x,x'\in \xi}\|x-x'\|$ and the \emph{diameter of $\cX$} by $d := \max_{\xi \in \cX} d (\xi)$. Note that for a uniform grid, $d = d(\xi)$, independently of the choice of $\xi$. Let $V(\xi)$ be the set of corner points of the cube $\xi$ and: 
\vspace{-.075in}
\[
\cF(\xi) := \left\{ \xi' \mid  \xi' \cap \overline{B\left(\phi_\tau(v),Ld/2 \right)} \neq \emptyset\ \text{for some $v\in V(\xi)$}  \right\}. 
\vspace{-.075in}
\] 

Fig. \ref{fig:outer_approx}(right) provides a relevant illustration. The definition of $\cF$ above is guaranteed to provide an outer approximation if $L$ is (an upper bound for) the Lipschitz constant  $L_\tau$. In practice, however, only an estimate for the Lipschitz constant $L_\tau$ is available. In this case, an outer approximation can be obtained by evaluating $\phi_\tau$ on a fine enough grid in $\xi$ instead of just the corner points.

\noindent {\bf Step 3 - Identification of recurrent \& non-recurrent behavior of $\cF$:} Identifying all the recurrent sets $\cM$ of $\cF$ is performed using Tarjan's Strongly Connected Components (SCC) algorithm, which is linear in the number of elements of $\cX$ plus the number of edges in $\cF$. The accompanying implementation uses a modified version of the algorithm, which does not store the whole digraph $\cF$ in memory and yet evaluates $\cF$ only once for each node \cite{Bush:Gameiro:Harker,CMGDB}.

An indexing set $\sP$ is introduced in order to distinguish all the recurrent sets and helps to enumerate them: $\{\cM(p)\mid p\in \sP\}$. A partial order relation $\leq$ on $\sP$ is imposed on the corresponding recurrent sets $\cM(p)$. In particular, $q \leq p$ if there exists a path in $\cF$ from a $\xi \in \cM(p)$ to a $\xi' \in \cM(q)$. Identifying the partial order $\leq$ for two recurrent sets is a question of reachability on $\cF$ between the two recurrent sets and is done taking advantage of the fact that Tarjan's algorithm also performs a topological sort on the vertices. For the examples of this paper, the number of recurrent sets is in the order of tens.

\begin{figure}[h]
    \vspace{-.25in}
    \centering
    \includegraphics[width=0.9\columnwidth]{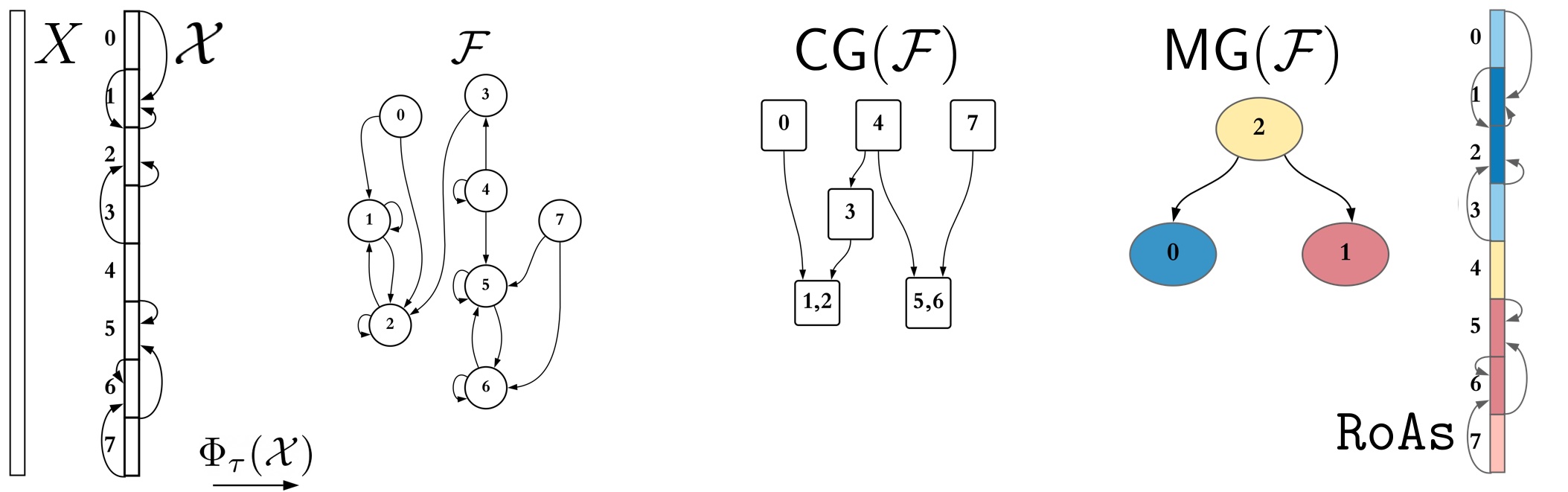}
    \vspace{-.15in}
    \caption{\small From left to right: An example decomposition of a state space into regions and transitions between regions. The corresponding multivalued map $\cF$. The condensation graph $\sCG(\cF)$ where SCCs have been condensed to a single vertex. The corresponding Morse graph were node 0 corresponds to the SCC \{1,2\} with RoA=\{0,1,2,3\} and node 1 corresponds to the SCC \{5,6\} with RoA=\{5,6,7\}. Initial conditions in region 4 (identified with node 2) may end up either in regions \{1,2\} (node 0) or \{5,6\} (node 1).}
    \label{fig:my_label}
    \vspace{-.3in}
\end{figure}

Thus, the output of the SCC algorithm is a new graph representation, a condensation graph $\sCG(\cF)$ of $\cF$, which is formed by contracting each strongly connected component of $\cF$ into a single vertex. The condensation graph $\sCG(\cF)$ is by definition a directed acyclic graph. The reachability on $\cF$ defines the direction of the edges in the condensation graph $\sCG(\cF)$ and relates to the partial order relation, where $v \leq w$, if there is a directed edge $w \to v$ in $\sCG(\cF)$. While the graph $\sCG(\cF)$ has condensed all SCCs into a single vertex, it is still a huge graph representation as it stores all the vertices of $\cF$ that are not in a recurrent set. 

For this reason, the proposed method outputs the sub-graph derived only from the recurrent sets (i.e., the non-trivial SCCs) as these components are the only possible candidates for containing the attractors of RoAs given the level of discretization. This is referred to as the \emph{Morse graph} $\sMG(\cF)$ of $\cF\colon \cX\mvmap \cX$ (shown in Fig.~\ref{fig:pendulum12}) and is the partially ordered set:
\vspace{-.1in}
\begin{equation}
\label{eq:MorseDec}
\sMG(\cF) = \{\cM(p)\subset \cX\mid p\in (\sP,\leq)\}.
\vspace{-.1in}
\end{equation}
Since $(\sP,\leq)$ is a partially ordered set, $\sMG(\cF)$ can be represented as a directed graph.  The \emph{Morse graph} $\sMG(\cF)$ of $\cF\colon \cX\mvmap \cX$ is the Hasse diagram of $(\sP,\leq)$, i.e., the minimal directed graph from which $(\sP,\leq)$ can be reconstructed. 

The Morse graph for the inverted pendulum is presented in Fig.~\ref{fig:pendulum12} and is indexed by $\sP = \{ 0, \ldots, 6 \}$ with order relations: $p < q$, iff there is a path from $q$ to $p$ in the digraph. Hence, $4$, $2$, and $0$ are the minimal elements and $1 < 3$ and $5 < 6$. The $\sMG(\cF)$ is computed by Algorithm~\ref{alg:MorseGraph}, which takes as input the decomposition $\cX$, the dataset $\Phi_\tau(\cX)$ (representing the map $\phi_\tau$), and an estimate for the Lipschitz constant $L$ of $\phi_\tau$.
The dataset $\Phi_\tau(\cX)$ is used to compute the outer approximation $\cF$, as described in Step 2.
When the condensation graph is computed, the non-trivial SCCs (components with at least one edge) are flagged, as they become nodes of the Morse graph. Then, the only step remaining is to determine the reachability of recurrent sets as discussed above.


\begin{algorithm}[H]
\small
\DontPrintSemicolon
$\cF \gets \FuncSty{OuterApproximation}(\cX, \Phi_\tau(\cX), L)$  \tcp*[h]{$\cF$ as a digraph but not stored in memory explicitly}\;
$\mathrm{SCC}(\cF) \gets \FuncSty{StronglyConnectedComponents}(\cF)$\;
$\mathrm{CG}(\cF) \gets \FuncSty{CondensationGraph}(\mathrm{SCC}(\cF))$  \tcp*[h]{flag non-trivial SCCs}\;
$\sMG(\cF) \gets \FuncSty{Reachability}(\mathrm{CG}(\cF))$\;
\KwRet{$\sMG(\cF), \mathrm{CG}(\cF)$}
\caption{\FuncSty{MorseGraph}($\cX$, $\Phi_\tau(\cX)$, $L$)}
\label{alg:MorseGraph}
\end{algorithm}



\noindent {\bf Step 4 - Derivation of RoAs:} 
Define $O_\bullet\colon \cX \mvmap \sP$ and $O^\bullet\colon \cX \mvmap \sP$ as:\\
$O_\bullet(\xi) := \min \{p\in \sP \mid \text{there exists a path in}~ \cF ~\text{from}~ \xi ~\text{to}~ \xi' \in \cM(p)\}$ and\\
$O^\bullet(\xi) := \max \{p\in \sP \mid \text{there exists a path in}~ \cF ~\text{from}~ \xi ~\text{to}~ \xi' \in \cM(p)\}$.

Note that since $\sP$ is a poset it is possible that $O_\bullet(\xi)$ and $O^\bullet(\xi)$ have multiple values. Under the assumption that $\cF$ is an outer approximation of $\phi_\tau$, then, if $p \not\in O_\bullet(\xi)$, it is true that for every $x\in \xi$ and any $n\geq 0$, $\phi_\tau^n(x)\not\in \cM(p)$.
\vspace{-.1in}


\begin{theorem}
\label{thrm:RoA}
If $p$ is a minimal element of $(\sP,\leq)$ and $O^\bullet(\xi) = \{ p \}$, then for every $x\in \xi$, there exists $n\geq 0$ such that $\phi_\tau^n(x)\in \cM(p)$.
As a consequence if $p$ is a minimal element of $(\sP,\leq)$, then $\left\{ \xi\in \cX \mid O^\bullet(\xi) = \{ p \} \right\}$ is the maximal RoA for $\cM(p)$ that can be rigorously identified using $\cF$.
\vspace{-.1in}
\end{theorem}

Returning to the example of Fig.~\ref{fig:pendulum12}, $O^\bullet(\xi) = p$ for $\xi$ is the region corresponding to the Morse set $\cM(p)$, for $p=0, \ldots, 6$. For $\xi$ in the RoA indicated by $a$, $b$, $c$, $d$, and $e$ the following map arises:
$O^\bullet(\xi) = 0$ for $a$, $O^\bullet(\xi) = 2$ for $b$, $O^\bullet(\xi) = 4$ for $c$, $O^\bullet(\xi) = 3$ for $d$, and $O^\bullet(\xi) = 6$ for $e$. This is indicated by the graph in Fig.~\ref{fig:pendulum12}(right). $O^\bullet(\xi)$ is the Morse node reachable from the corresponding region in the graph. It follows from Theorem~\ref{thrm:RoA} that the regions in Fig.~\ref{fig:pendulum12} labeled $4$ and $c$ form the maximal RoA of $4$, the regions labeled $2$ and $b$ form the maximal RoA of $2$, and the regions labeled $0$ and $a$ form the maximal RoA of $0$.

To obtain $O^\bullet$, the graph $\sCG(\cF)$ is explored with a depth first search (DFS) approach and for each visited vertex $v \in \sCG(\cF)$ the maximal reachable Morse nodes are identified. That is, the collection of $p\in \sP$ such that there exists a path from $v$ to a cube $\xi\in\cM(p)$. See Appendix \ref{sec:appendix-algorithm} and \cite{RoA} for an implementation of the DFS.



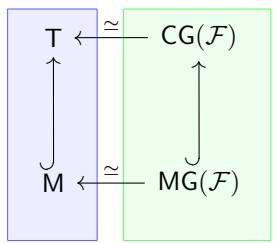
\begin{wrapfigure}{l}{0.4\textwidth}
\vspace{-.35in}
\centering
\begin{tikzpicture}[scale=1.4]
\node[] at (0.15,1.8) {{\color{blue} Dynamics}};
\node[] at (1.6,1.8) {{\color{green}   Combinatorics}};
\draw[green!60!white]  (0.9,-0.6) to (0.9,1.6);
\draw[green!60!white] (2.3,-0.6) to (2.3,1.6);
\draw[green!60!white]  (0.9,-0.6) to (2.3,-0.6);
\draw[green!60!white] (0.9,1.6) to (2.3,1.6);
\draw[blue!60!white] (0.65,-0.6) to (0.65,1.6);
\draw[blue!60!white] (-0.2,-0.6) to (-0.2,1.6);
\draw[blue!60!white] (-0.2,-0.6) to (0.65,-0.6);
\draw[blue!60!white](-0.2,1.6) to (0.65,1.6);
\filldraw[blue!70!white,opacity=0.1] (-0.2,-0.6) to (-0.2,1.6) to (0.65,1.6) to (0.65,-0.6) to (-0.2,-0.6);
\filldraw[green!70!white,opacity=0.1] (2.3,-0.6) to (2.3,1.6) to (0.9,1.6) to (0.9,-0.6) to (2.3,-0.6);
$
\begin{diagram}
\node{\sT} 
\node{\sCG(\cF)}  \arrow{w,l}{\simeq} \\
\node{\sM}  \arrow{n,l,J}{} 
\node{\sMG(\cF)} \arrow{n,l,J}{} \arrow{w,l}{\simeq}  \\
\end{diagram}
$
\end{tikzpicture}
\caption{\small Diagram relating the discretized \emph{dynamics} in state space (left) to the proposed \emph{combinatorial representation} of the global dynamics (right). The arrows $\xhookrightarrow{}$ represent inclusion maps and the arrows $\xrightarrow{\simeq}$ indicate poset isomorphisms.}
\label{diag:framework}
\vspace{-.35in}
\end{wrapfigure}

\noindent {\bf Relationship to Continuous Dynamics:}  The \emph{condensation graph} $\sCG(\cF)$ and the \emph{Morse graph} $\sMG(\cF)$ of $\cF$ are highlighted in the right column of Fig.~\ref{diag:framework} as the combinatorial objects computed  given the outer approximation $\cF$ of $\phi_\tau$. Each element of $\sCG(\cF)$ is identified with a region of $X$. This collection of regions is denoted by $\sT$ and, as shown in Fig~\ref{diag:framework}, it is isomorphic as a poset to $\sCG(\cF)$. To find attracting blocks, the method uses the following fact \cite{kalies:mischaikow:vandervorst:15}. Let $\cF$ be an \emph{outer approximation} of $\phi_\tau$: If $x\in T\in \sT$ and $\phi_\tau(x)\in T'\in \sT$, then $T'\leq T$ where $\leq$ is the order relation on $\sT$, i.e., forward orbits under $\phi_\tau$ can be tracked by descending the order relation on $\sT$. 
Stated more concisely: $\sO(T)\in \sABlock(\phi_\tau)$, where
$\sO(T):= \{T'\in \sT \mid T'\leq T\}$ is the  \emph{downset} of $T\in\sT$.

The set $\sABlock(\phi_\tau)$ has the structure of a finite distributive lattice; that is, if $N, N'\in\sABlock(\phi_\tau)$ then $N\cup N', N\cap N'\in \sABlock(\phi_\tau)$.
Furthermore, the collection $\{\sO(T) \mid T\in \sT \}$ generates a finite but huge sublattice of $\sABlock(\phi_\tau)$. The elements of $\sT$ that are identified by elements of $\sMG(\cF)$ are denoted by $\sM$ and referred to as \emph{Morse cells} (see Fig.~\ref{fig:pendulum12}).
Given prior work \cite{kalies:mischaikow:vandervorst:15}, if $\cF$ is an outer approximation of $\phi_\tau$ and $x\in X$ belongs to the chain recurrent set of $\phi_\tau$ (recurrence allowing for an arbitrarily small error \cite{conley:cbms}), then $x$ belongs to a Morse cell. Thus, the collection of recurrent sets of $\cF$ identifies the location in state space of recurrent dynamics of $\phi_\tau$.
As in Fig.~\ref{diag:framework}, $\sM$ inherits a partial order $\leq$ from $\sMG(\cF)$. Its dynamical implications are derived from the dynamical implications of the partial order on $\sT$. If $M,M'\in \sM$, $M < M'$  and $x$ is an initial condition that lies in $M$, then $\phi_\tau^n(x)\cap M' = \emptyset$ for all $n \geq 0$.
\vspace{-.15in}

\section{Results}
\label{sec:results}
\vspace{-.05in}

The framework is compared against alternatives for estimating the RoAs, including numerical and machine learning methods that compute Lyapunov functions.

\textbf{Robotic Systems:} (i) A 2$^{nd}$-order (\texttt{Pendulum}), (ii) a 1$^{st}$-order car with (\texttt{Ackermann}) steering that cannot reverse \cite{corke2011robotics}, (iii) a $2^{nd}$-order (\texttt{Acrobot}) \cite{spong_acrobot}. The dynamics are simulated via numerical integration \cite{ML4KP}. The state and control space limits are given in Table~\ref{table:systems}. 

\vspace{-.25in}
\begin{table}[]
\centering
\begin{tabular}{|c|c|c|c|c|c|c|}
\hline
\textbf{System} & \textbf{$X$} & $\mathbb{U}$ & \textbf{Bounds on $X$}  & \textbf{Bounds on $\mathbb{U}$} & \textbf{Controllers} & \textbf{Goal} \\ \hline
{\scriptsize\texttt{Pendulum} } & ${\scriptstyle(\theta, \dot{\theta})}$ & $\tau$ &$ {\scriptstyle[\{-\pi,-2\pi\},\{\pi,2\pi\}]}$ & ${\scriptstyle[0.6372,0.6372]}$ & {\scriptsize\texttt{Learned, LQR} }& ${\scriptstyle[0,0]}$  \\ \hline
{\scriptsize\texttt{Ackermann} }& ${\scriptstyle(x,y,\theta)}$ & ${\scriptstyle(\gamma, V )}$ & ${\scriptstyle[\{-10,-10,-\pi\},\{10,10,\pi\}]}$  & ${\scriptstyle[\{-\frac{\pi}{3},0\},\{\frac{\pi}{3},30\}]}$ & \makecell{{{\scriptsize\tt Learned, LQR}}, \\{{\scriptsize\tt Corke}}} & ${\scriptstyle[0,0,\frac{\pi}{2}]}$ \\ \hline
{\scriptsize\texttt{Acrobot} }  & ${\scriptstyle(\theta_1, \theta_2, \dot{\theta_1}, \dot{\theta_2})}$ & $\tau_2$ & ${\scriptstyle[\{0,-\pi,-6,-6\},\{2\pi,\pi,6,6\}]}$ & ${\scriptstyle[-14,14]}$ & {\scriptsize\texttt{Hybrid, LQR}} & ${\scriptstyle[0,\pi,0,0]}$  \\ \hline
\end{tabular}

\caption{\small Systems and controllers considered in the evaluation.}
\label{table:systems}
\vspace{-.45in}
\end{table}

\textbf{Controllers:} For each system, alternative controllers are considered:\\
(i) An \texttt{LQR} controller linearizes the system around the goal: $\dot{x} = Ax + Bu\ (A \in \mathbb{R}^{n \times n}, B \in \mathbb{R}^{m \times n})$
and the controller $u = -Kx$ minimizes the cost $\mathcal{J}_{LQR} = x(T)^TFx(T) + \int_{0}^T (x(t)^TQx(t) + u(t)^TRu(t)) dt$ ($F, Q \in \mathbb{R}^{n \times n}, R \in \mathbb{R}^{m \times m}$).\\
(ii)  A \texttt{Learned} controller trained using the Soft Actor-Critic (SAC) \cite{Haarnoja2018SoftAO} algorithm to maximize the expected return $\mathcal{J}(\pi) = \mathbb{E}_{\tau \sim \rho_\pi} [\sum_{t=0}^T \mathcal{R}(x_t)]$, where the reward function is $\mathcal{R}: X \rightarrow \{0,1\}$. $\mathcal{R}(x_t) = 1$ iff  $x_t$ is within an $\epsilon$ distance from the goal state and 0 otherwise.\\
(iii) A {\tt Hybrid} controller for the \texttt{Acrobot} uses the {\tt Learned} controller to drive the system to a relaxed goal distance, from where {\tt LQR} takes over.\\
(iv) The \texttt{Corke} controller for \texttt{Ackermann} \cite{corke2011robotics} transforms the state into polar coordinates $(\rho, \alpha, \beta)$ to define linear control laws for the velocity $v=K_{\rho}\rho$ and steering angle $\omega= K_{\alpha}\alpha + K_{\beta}\beta$. It's able to drive a reverse-capable system to the goal assuming $K_{\rho} > 0, K_{\beta} < 0, K_{\alpha} - K_{\rho} > 0$. The experiments, however, consider only positive velocities limiting its reachability.






\textbf{Comparison Methods:} \texttt{L-LQR} and \texttt{L-SoS} use a linearized, unconstrained form of the dynamics to compute a Lyapunov function (LF) for the controller being considered. \texttt{L-LQR} uses the solution of the Lyapunov equation $v_{LQR}(x)=x^TPx$ for the linearized, unconstrained version of the system. \texttt{L-SoS} computes the LF according to $v_{sos}(x)=m(x)^TQm(x)$ where $m(x)$ are monomials on $x$ and $Q$ is a positive semidefinite matrix. It uses SOSTOOLS \cite{prajna2002introducing} with SeDuMi \cite{sturm1999using} as the SDP solver. These methods cannot be used with data-driven controllers (like {\tt Learned}) since they require a closed-form expression for the controller. 

The Lyapunov Neural Network ({\tt L-NN}) \cite{richards2018lyapunov} is a state-of-the-art, machine learning approach with available software for identifying RoAs of black-box controllers. It returns a parametrized function that is trained to adapt to the RoA of a closed-loop dynamical system. Given an initial safe set around the desired equilibrium, a subset of non-safe states are forward propagated, classified and used to reshape the Lyapunov candidate in each iteration. The method needs access to a known safe-set and there is no guarantee the safe-region won't shrink.

\begin{wraptable}{r}{0.6\textwidth}
    \centering
    \vspace{-.35in}
    \small
    \begin{tabular}{|c||c|c|c|c|}
        \hline
        \textbf{Benchmark} & {\tt L-NN} & {\tt L-LQR} & {\tt L-SOS} & Ours: {\tt MG}    \\ \hline \hline
        Pend (LQR) & \textbf{97.54\%} & 69.91\% & 3.07\%  & 97.49\% \\ \hline 
        Pend (Learned) & 30.18\% & \cellcolor{gray!} & \cellcolor{gray!} & \textbf{98.5\%} \\ \hline
        Acro (LQR) & 89.06\% & 26.84\% & 25.66\% & \textbf{96.36\%} \\ \hline
        Acro (Hybrid) & 13.79\% & \cellcolor{gray!} & \cellcolor{gray!} & \textbf{98.75\%} \\ \hline
        Ack (LQR) & 7.55\%  & \textbf{21.78\%}  & 2.43\% & 0\%  \\ \hline
        Ack (Corke) & 10.23\%  & \cellcolor{gray!} & 41.36\% & \textbf{86.69\%} \\ \hline
        Ack (Learned) & 91.47\% & \cellcolor{gray!} & \cellcolor{gray!} & \textbf{100\%} \\ \hline 
    \end{tabular}
    \vspace{-.1in}
    \caption{\small RoA ratios identified by the different methods. Best values per row in bold.}
    \label{tab:tp}
    \vspace{-.40in}
\end{wraptable}

\textbf{``Ground Truth'' RoAs:} For each benchmark (i.e., a controller-system pair), an approximation of the ground truth RoA for the goal state is computed by high-resolution discretization of the state space, and forward propagating the controller for a very long, fixed time horizon, or until the goal is reached. Appendix \ref{sec:appendix-experiments} provides the parameters for this ground truth evaluation.

\noindent  \textbf{Metrics:} Given the ``ground truth'' RoA, the following metrics are reported:\\ (a) Table~\ref{tab:tp} provides the ratio of $X$'s volume correctly identified to belong to the RoA (True Positives - TP) - its complement gives the ratio of $X$'s volume \textit{incorrectly} identified as not being in the RoA (False Negatives - FN);\\ (b) Table~\ref{tab:tn} provides the ratio of $X$'s volume for which the dynamics have not been identified (Unidentified);\\ (c)  Table~\ref{tab:steps} provides the amount of computational resources required for {\tt L-NN} and {\tt Morse Graph} ({\tt MG}) measured using the number of forward propagations performed (steps) -- the dominant computational primitive. Given their analytical nature, the {\tt L-LQR} and {\tt L-SOS} alternatives do not require forward propagations of the system and tend to be computationally faster but require access to an expression for the controller.


\begin{wraptable}{r}{0.6\textwidth}
    \centering
    \vspace{-.35in}
    \small
    \begin{tabular}{|c||c|c|c|c|}
        \hline
        \textbf{Benchmark} & {\tt L-NN} & {\tt L-LQR} & {\tt L-SOS} & Ours: {\tt MG} \\ \hline \hline
        Pend (LQR) & 61.33\%  & 61.33\%  & 61.33\%    & \textbf{1.66\%} \\ \hline 
        Pend (Learned) & 81.44\%  & \cellcolor{gray!} & \cellcolor{gray!} & \textbf{1.25\%} \\ \hline
        Acro (LQR) & 10.94\% & 73.16\% & 74.34\% & \textbf{3.64\%} \\ \hline
        Acro (Hybrid) & 86.21\% & \cellcolor{gray!} & \cellcolor{gray!} & \textbf{1.25\%} \\ \hline
        Ack (LQR) & 83.03\%  & \textbf{82.87\%}   & \textbf{82.87\%}  & 100\%   \\ \hline
        Ack (Corke) & 27.81\%   & \cellcolor{gray!} & 29.35\%  & \textbf{24.5\%}  \\ \hline
        Ack (Learned) & 8.53\% & \cellcolor{gray!} & \cellcolor{gray!} & \textbf{0.00\%}\\ \hline 
    \end{tabular}
    \vspace{-.15in}
    \caption{\small $X$'s ratio returned as \textit{unidentified}
by the different methods. Best values per row in bold.}
    \label{tab:tn}
    \vspace{-.35in}
\end{wraptable}

\noindent \textbf{Quantitative Results:} The proposed method tends to estimate larger volumes of the RoA compared to alternatives per Table \ref{tab:tp}. It also consistently identifies the dynamics for larger volumes of $X$ compared to the comparison points, which cover lower volumes of $X$ per Table \ref{tab:tn}.  Moreover, the proposed method is broadly applicable to different controllers, and finds a larger volume of the RoA when compared to {\tt L-NN} for the Pendulum (Learned), Acrobot (Hybrid), Ackermann (Learned). 


\begin{wraptable}{l}{0.45\textwidth}
    \centering
    \vspace{-.35in}
    \small
    \begin{tabular}{|c||c|c|}
        \hline
        \textbf{Benchmark} & {\tt L-NN} & Ours: {\tt MG}    \\ \hline \hline
        Pend (LQR) & 667.1M & \textbf{6.6M} \\ \hline 
        Pend (Learned) & 341.9M & \textbf{6.6M} \\ \hline
        Acro (LQR) & 5.7B  &  \textbf{1.1B} \\ \hline
        Acro (Hybrid) & \textbf{533M} & 2.1B  \\ \hline
        Ack (LQR) & \textbf{9.9M}  & 520M \\ \hline
        Ack (Corke) & 37.5M & \textbf{13M}  \\ \hline
        Ack (Learned) & 704.6M  & \textbf{520M}  \\ \hline 
    \end{tabular}
    \vspace{-.1in}
    \caption{\small Number of propagation required. Best values per row in bold.}
    \label{tab:steps}
    \vspace{-.35in}
\end{wraptable}

The computational needs of the Morse Graph is one or two orders of magnitude less than that of the {\tt L-NN}. The learned controllers benefit the most from the topological approach as it provides in all cases a higher coverage of the RoA with fewer propagations. There are two cases where {\tt MG} takes a larger number of propagation steps compared to {\tt L-NN}. In the case of Ackermann (LQR), this is because {\tt MG} is unable to find a unique attractor for the system (see discussion below). In the case of Acrobot (Hybrid), {\tt L-NN} fails to identify the true RoA accurately.



Some of the comparison points may incorrectly identify a volume of $X$ as belonging to RoA (False Positives - FP). This is not true for the Pendulum, since the attractor of interest is not at the boundary of the RoA. But for the LQR and Corke controllers of the Ackermann, the desired goal is not an attractor since some trajectories close to the goal region escape $X$ (a consequence of not allowing negative velocities). Therefore, the comparison methods fail to conservatively estimate the RoA, resulting in FPs. The topological framework, however, does not result in FPs (explained below). The Ackermann Learned controller and the Acrobot controllers present no FPs since the goal region is an attractor.

\begin{figure}[h!]
\vspace{-.2in}
\centering
\begin{overpic}[width=0.45\columnwidth]
{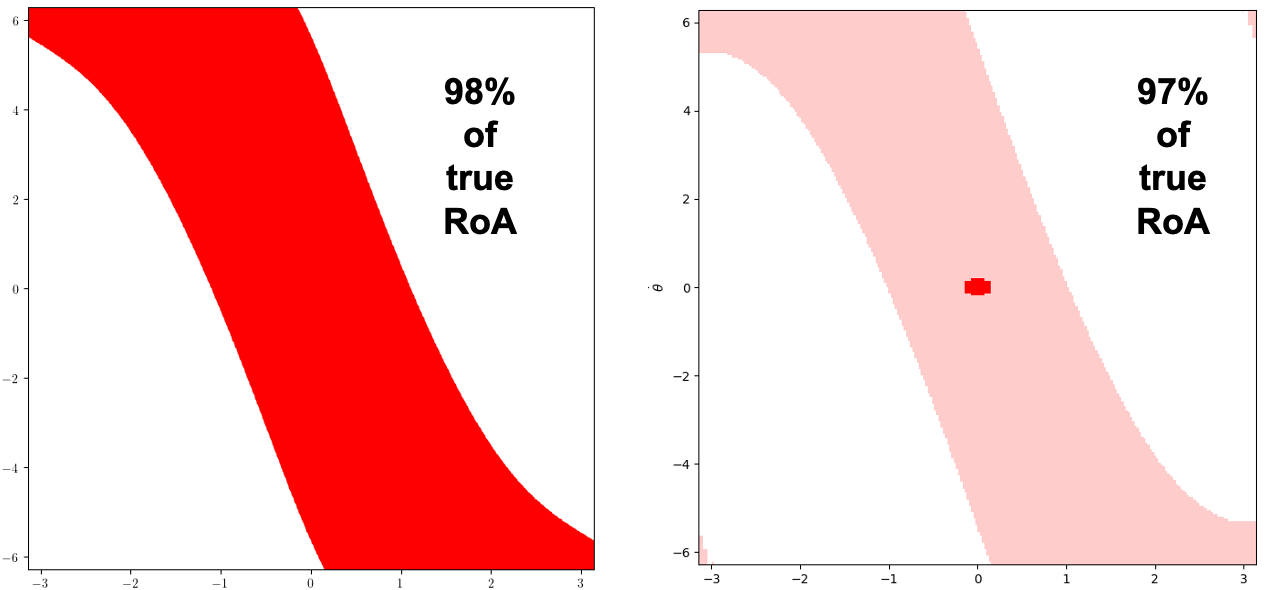}
\put(-4,22){\scriptsize $\dot\theta$}
\end{overpic}
\begin{overpic}[width=0.45\columnwidth]
{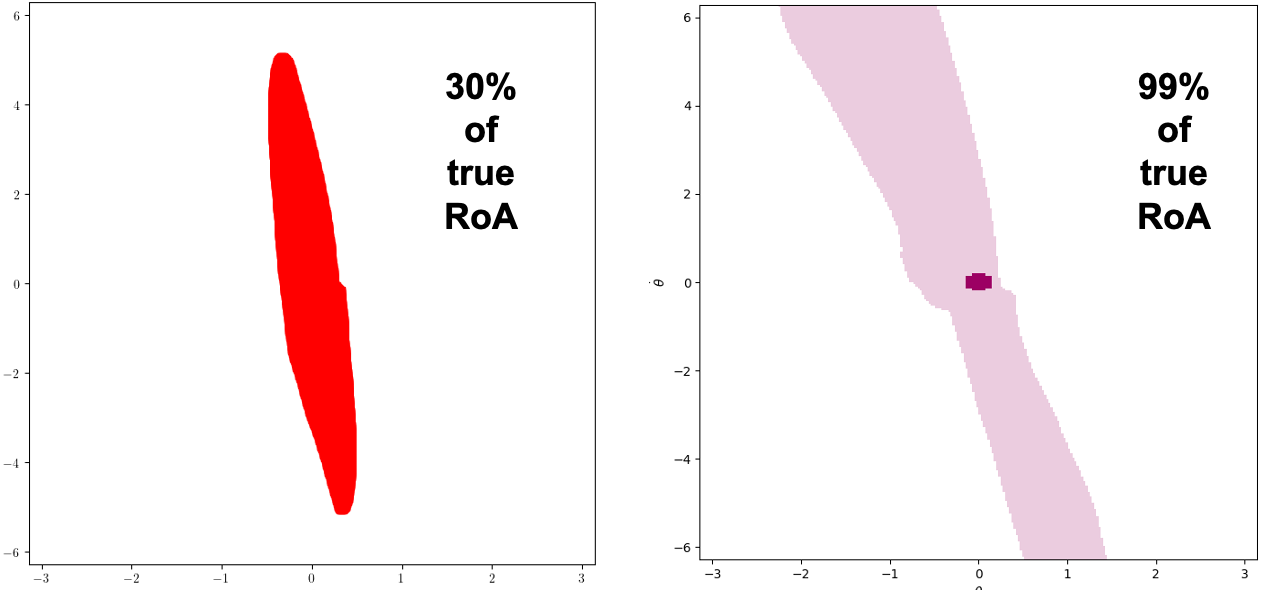}
\put(101,1){\scriptsize $\theta$}
\end{overpic}
\vspace{-.1in}
\caption{\small RoA estimated by {\tt L-NN} (1$^{st}$ and 3$^{th}$ image) and {\tt Morse Graph} (2$^{nd}$ and 4$^{th}$ image, where the dark shading is the attractor) for the LQR (left) and Learned controllers (right) of the pendulum.}
\label{fig:pendulum-qual}
\vspace{-.35in}
\end{figure}

\textbf{Pendulum Study:} The RoAs compute controllers for the pendulum are shown in Fig~\ref{fig:pendulum-qual}. For {\tt Morse Graph}, the attractor discovered is shown at the center of the state space, and it contains the goal region. Note that {\tt L-NN}  does not find the attractor, but assumes one exists containing the goal region. Moreover, {\tt L-NN} is unable to cover the full RoA for the Learned controller, making it an undesirable solution. {\tt Morse Graph} is also shown to work well when (a subset of) the state space is periodic. For instance, the RoA of the Pendulum (LQR) also includes the regions at the corners of the planar representation of the cylinders in Fig~\ref{fig:pendulum-qual} (Left). This is captured by {\tt MorseGraph}, but not by {\tt L-NN}.

\begin{figure}[h!]
\vspace{-.3in}
\centering
\begin{overpic}[width=0.9\textwidth]{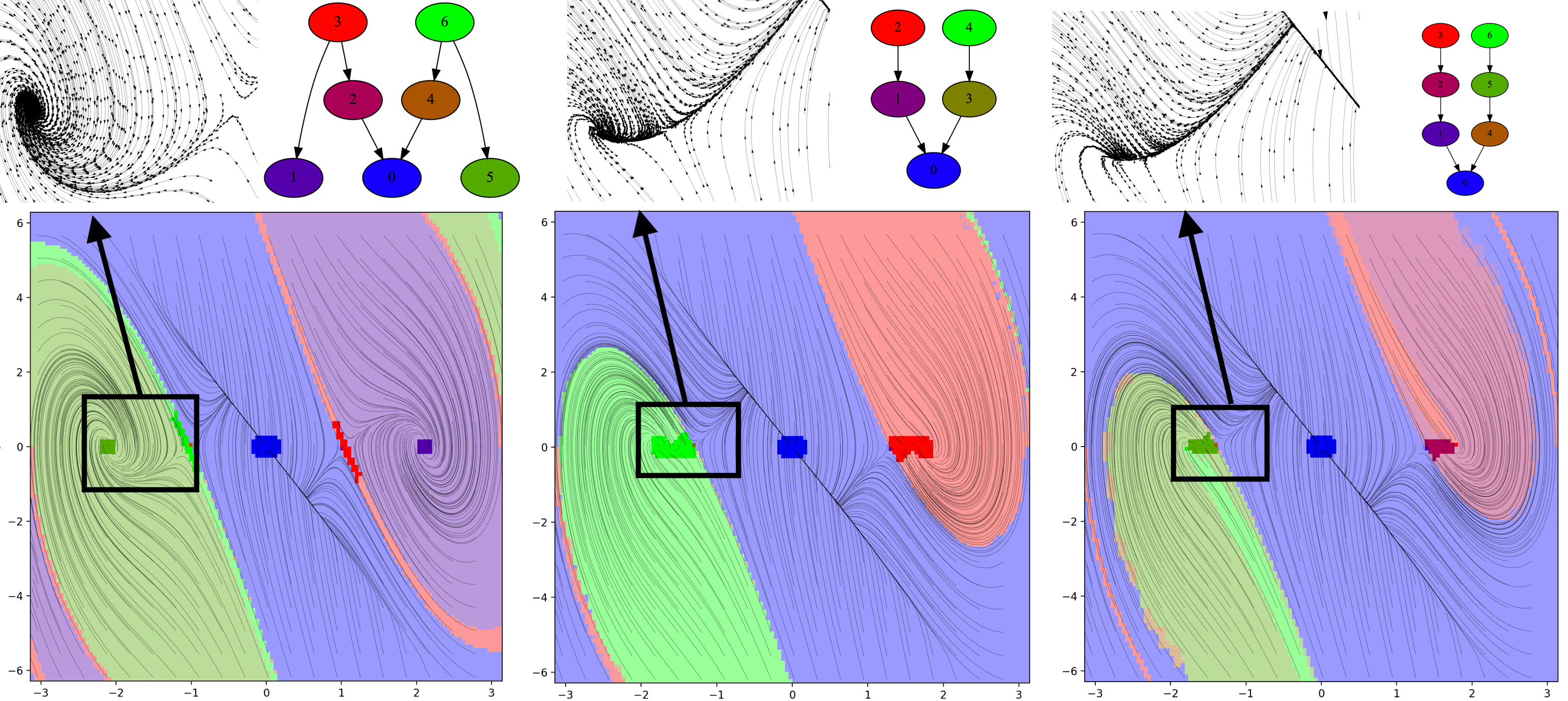}
\put(0,13){\scriptsize $\dot\theta$}
\put(66,1){\scriptsize $\theta$}
\end{overpic}
\vspace{-.1in}
\caption{\small RoA for the Pendulum when the upper torque bound is: (left) $0.637$; (center) $0.724$; (right) $0.736$. For all torques the RoA found is $97\%$ of the true one.}
\label{fig:pendulum-torques}
\vspace{-.35in}
\end{figure}

\emph{Pendulum with different torques:} Fig~\ref{fig:pendulum-torques} illustrates the robustness of the topological framework to different bounds for the allowed torque of the LQR controller. The proposed method consistently finds the RoA and the attractor that contains the goal region. Moreover, it persistently covers $99\%$ of the true RoA even when the dynamics  change, for instance in Fig. \ref{fig:pendulum-torques} one attractor and one saddle eventually collide and disappear.



\noindent \textbf{Ackermann Study: } \emph{Learned controller for Ackermann:} Although the RoA is the full state space, only the Morse Graph identifies this.
The unique attractor obtained (right) also provides insight into how the controller works. For a small resolution of the discretization of $X$, it presents as a torus-like shape, which suggests recurrent behavior. This shape can be explained by the behavior of the car when it gets close to the goal region with the wrong orientation, where it tries to fix this orientation by performing a loop. For a more refined discretization, the method can distinguish long trajectories from recurrent behavior, finding a smaller attractor that contains the goal region.

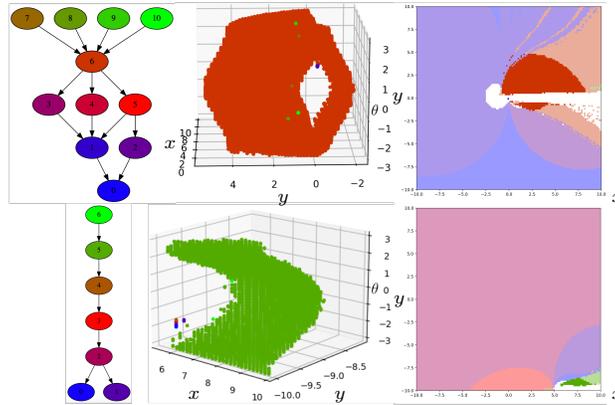
\begin{figure}[!h]
\centering
\vspace{-.25in}
\begin{overpic}[width=0.65\columnwidth]{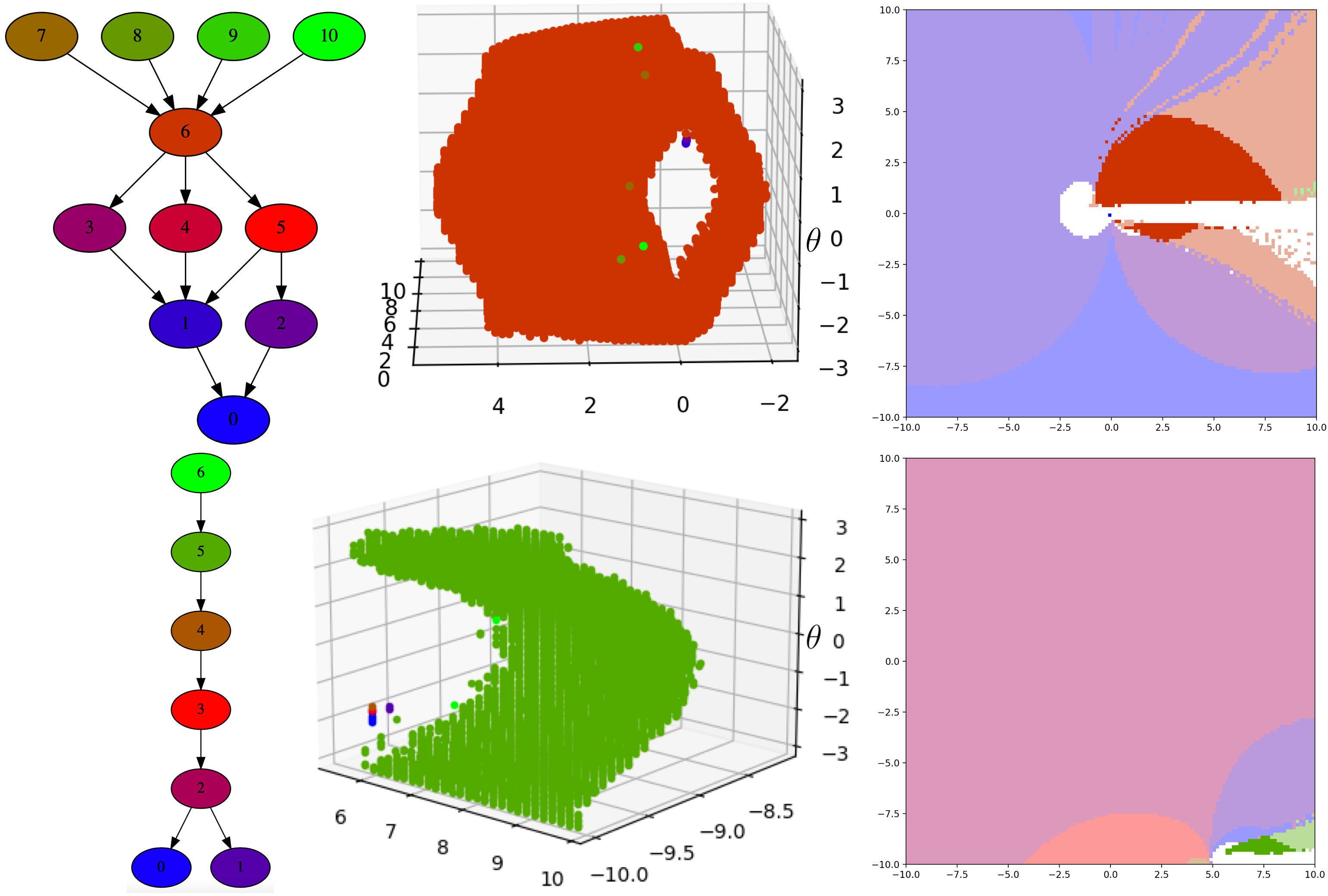}
\put(45,34){\scriptsize $y$}
\put(54.5,1){\scriptsize $y$}
\put(64.2,51){\scriptsize $y$}
\put(64.5,17){\scriptsize $y$}
\put(26,43){\scriptsize $x$}
\put(30,1){\scriptsize $x$}
\put(101,34){\scriptsize $x$}
\put(101,1){\scriptsize $x$}
\end{overpic}
\vspace{-.15in}
\caption{\small Morse Graph, Morse sets and 2D projection on $\theta = 0$ of the RoA for the Corke controller applied to the Ackermann. (top row) Corke controller with the goal is set to be $(0, 0, 1.57)$ and (bottom row) Corke controller with the goal set to be $(6, -10, -1.57)$.}
\label{fig:ackermann_sys2}
\vspace{-.25in}
\end{figure}

\emph{LQR controller for Ackermann:} The Morse Graph results in a large, unique attractor that contains $74\%$ of $X$ as the discretization is proven insufficient.  The comparison points fail by producing false positives (FP). The proposed method is conservative and safe. It avoids FPs but needs more subdivisions for a more comprehensive understanding of these global dynamics.\\
\begin{wrapfigure}{r}{0.35\textwidth}
    \centering
    \vspace{-.55in}
    \begin{overpic}[width=0.35\textwidth]{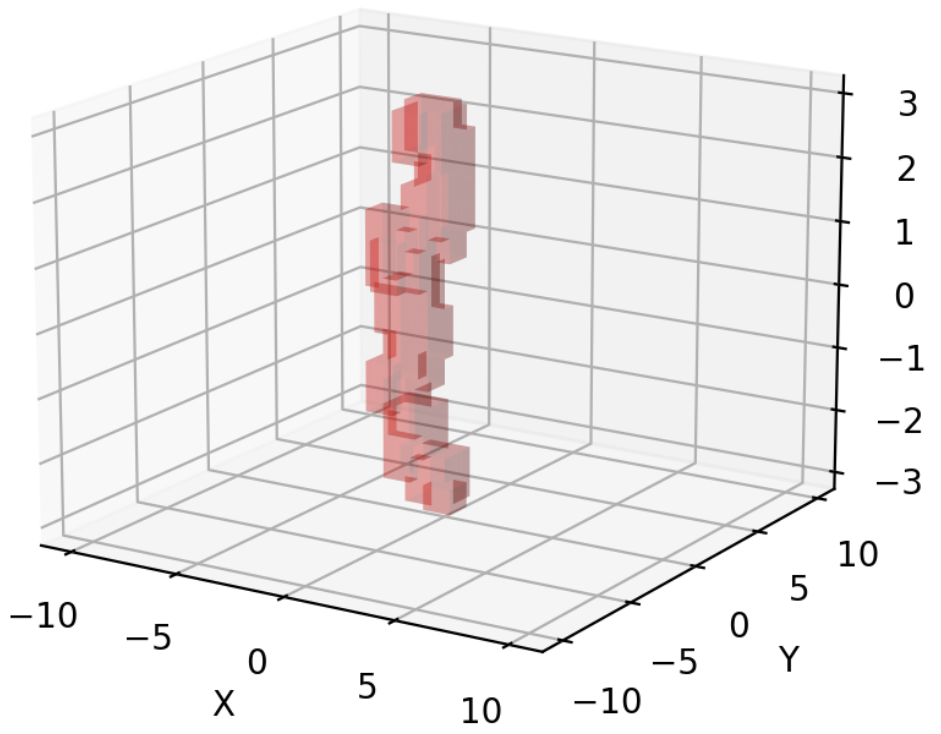}
    \put(98,45){\scriptsize $\theta$}
    \end{overpic}
    \vspace{-0.2in}
    \label{fig:ackermann_sys}
    \caption{\small For Ackermann (Learned), the entire state space is correctly identified by the Morse Graph as an RoA for the shown Unique Morse set (Torus-like shape).}
    \vspace{-0.35in}
\end{wrapfigure}
\emph{Corke controller for Ackermann:} Some trajectories leave $X$ in this experiment and modifications are needed to compute the RoA. A node $\star$ is included in $\sCG(\cF)$, and for every cube $\xi \in \cX$ s.t. $\Phi_{\tau}(\xi) \cap X^c \neq \emptyset$, edges are added from $\xi$ to $\star$ before computing the Morse Graph.  Thus, $\star$ is a minimal node of $\sMG(\cF)$. Then, a modified RoA computation is applied where $O^\bullet$ and $\max$ are changed to $O_\bullet$ and $\min$, respectively, and the resulting output is $O_\bullet$. Finally, for each element $R \in \smRoA$, the method computes $R_\bullet = R - R_\star$, where $R_\star = \{\xi \in \cX \ |\ O_\bullet(\xi) = \{\star\}\}$. Consequently, for each maximal RoA, the cubes in $R_\star$ are removed since they have some trajectories that escape $X$. So, $R_\bullet$ is a conservative estimate of the RoA. In Fig~\ref{fig:ackermann_sys2}, the white region is the set of cubes in $R_\star$.

\emph{Devising a Hybrid controller for Ackermann given the Morse Graph output: }
Given the information from the Morse Graph, it is possible to synthesize a hybrid controller that has a bigger RoA than the original ROA$_\text{Corke}$ of the Corke Controller $u_\text{init}$. The strategy selects a state in ROA$_\text{Corke}$, different than the original goal, as the goal for a new Corke controller $u_\text{inter}$. Define as  RoA$_{\text{inter}}$ the RoA of the new Corke controller. If RoA$_{\text{inter}}$ overlaps with $X - $RoA$_{\text{Corke}}$, the hybrid controller is defined as: for states in ROA$_\text{Corke}$, apply $u_\text{init}$, and for states in $X - $RoA$_{\text{Corke}}$, apply $u_\text{inter}$, and then apply $u_\text{init}$ when the system enters ROA$_\text{Corke}$. Fig~\ref{fig:ackermann_sys2} (bottom) shows the RoA for $u_\text{inter}$ with goal (6, -10, -1.57), which is in the ROA$_\text{Corke}$. A new controller $u_\text{inter}$ was devised for this goal and its RoA$_{\text{inter}}$ contained $X -$RoA$_{\text{Corke}}$. The integration of $u_\text{inter}$  with the original $u_\text{init}$ result in a hybrid solution that covers the entire state space, which was verified empirically using the Morse Graph.

    
    
    


\begin{wrapfigure}{r}{0.07\textwidth}
    \centering
    \includegraphics[width=0.06\textwidth]{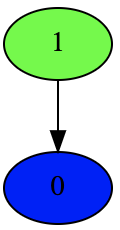}
    \vspace{-0.1in}
    \caption{}
    \label{fig:2_node_MG}
    \vspace{-0.35in}
\end{wrapfigure}
\textbf{Acrobot study:}  The Morse graph for both controllers of the Acrobot is shown in Fig~\ref{fig:2_node_MG}. They have a high success rate: the estimated RoA for node 0 covers more than $93\%$ of the true RoA. The possible recurrent dynamics described by node $1$ are long trajectories interpreted as recurrent by the proposed method. This can be addressed by increasing the state space discretization for an additional computational cost.


\emph{Hybrid Controller for Acrobot:}
The Acrobot learned controller is unable to find a solution within a given time horizon for a goal condition: $\mathcal{B}(x_G,0.1)$. Hence, a hybrid solution is proposed. The learned controller is first applied until the system reaches a relaxed goal: $\mathcal{B}(x_G,0.6)$.  The proposed method finds that the RoA for the relaxed goal condition is 100\% of the state space, and is inside the RoA of the LQR controller. Hence, once the trajectory reaches $\mathcal{B}(x_G,0.6)$ using the learned controller, LQR is applied to get into the smaller region $\mathcal{B}(x_G,0.1)$. This results in a hybrid controller with a 100\% success rate and a reduction in trajectory length relatively to just executing LQR of around 50\%.

\vspace{-.15in}
\section{Conclusion}
\label{sec:conclusion}
\vspace{-.1in}

This work present a novel method based on topology to identify attractors and their RoAs for robotic systems controlled by black-box controllers. Experimental evaluation on simulated benchmarks shows that the proposed method efficiently identifies the global dynamics with fewer samples from the dynamics model compared to data-driven alternatives. It does not require knowledge of the system or controller dynamics, such as differentiability or the guaranteed presence of an attractor for the system's goal region. This makes it suitable for data-driven controllers, where it significantly outperforms alternatives in identifying their RoA. Moreover, the proposed method provides a compact description of the global dynamics, which allows to compose multiple controllers into hybrid solutions that reach the goal from the full state space. The evaluation section presents two such hybrid controllers designed based on the Morse Graph output that yield notable properties: one increased the RoA to the whole state space; and the other decreased the length of solution trajectories by half.

Even though the proposed method requires less number of samples from the dynamics model, it still explores the entirety of the state space. This can potentially be mitigating for large and high-dimensional state spaces. Future work will explore extensions of the current topological approach, where the focus will be on finding the RoA for a single attractor, thereby requiring less computational resources. Integration with Gaussian Processes and machine learning primitives can help identify a smaller set of states where the system is propagated from so as to further reduce data requirements.
\vspace{-.15in}

\bibliographystyle{splncs04}
\bibliography{refs.bib}

\appendix

\section{Supportive Information for the Computation of RoAs}
\label{sec:appendix-algorithm}

This section presents the pseudocode for computing RoAs. It consists in a modified version of depth first search to explore the condensation graph $\sCG(\cF)$, where it assigns the maximal Morse nodes to each vertex of $\sMG(\cF)$. More specifically, the Algorithm \ref{alg:RoA} in lines 2-4, assigns the index of $w$ in $\sMG(\cF)$, named $p$, for each vertex $w$ in $\sMG(\cF)$,  to $O^\bullet(w)$. Note that $p$ is a maximal Morse node for $w$. Line 8 calls the function \texttt{propagate} described by Algorithm \ref{alg:propagate}, where it explores recursively the adjacency of each vertex in $G$ (set of vertices in $\sCG(\cF)$ that are not in $\sMG(\cF)$). After the recursive call in line 5, the map $O^\bullet$ is updated, in line 6, with the maximal Morse nodes found so far, see \cite{RoA} for an implementation.

\begin{algorithm}[H]
\small
\DontPrintSemicolon
$O^\bullet \gets \emptyset$  \tcp*[h]{set $O^\bullet(v) \gets \emptyset$ for all vertices $v$ of $\sCG(\cF)$}\;
\For{$w\in \sMG$}{
$p \gets \sP$-index of $w$  \tcp*[h]{index $p \in (\sP,\leq)$ such that $w=M(p)$}\;
$O^\bullet(w) \gets \{p\}$\;
}
$G \gets ~\text{vertices of}~ \sCG(\cF) \setminus \text{vertices of}~ \sMG(\cF)$\;
\While{$G \neq \emptyset$} {
$v \gets G.pop()$\;
$\FuncSty{propagate}(v, \sCG(\cF).adjacencies(v), O^\bullet)$
}
\KwRet{$O^\bullet$}
\caption{\FuncSty{RegionsOfAttraction}$(\sMG(\cF), \mathrm{CG}(\cF))$}
\label{alg:RoA}
\end{algorithm}

\begin{algorithm}[H]
\small
\DontPrintSemicolon
\For{$u \in A$} {
    \eIf{$O^\bullet(u)\neq \emptyset$}{
      $O^\bullet(v) \gets \max ( O^\bullet(v) \cup O^\bullet(u) )$\;
    }{
    $\FuncSty{propagate}(u, \sCG(\cF).adjacencies(u), O^\bullet)$\;
    $O^\bullet(v) \gets \max ( O^\bullet(v) \cup O^\bullet(u) )$\;
    $G \gets G \setminus \{u\}$\;
    }
}
\caption{\FuncSty{propagate}$(v, A, O^\bullet)$}
\label{alg:propagate}
\end{algorithm}

\section{Discretization of the State Space and the Time}

Two parameters play an integral role in identifying attractors and RoAs: the resolution of the discretization (R) and the forward propagation time of the dynamics (H). Without \textit{a priori} knowledge of the dynamics, it is not possible to theoretically estimate the optimal values for both the parameters. Nevertheless, a heuristic approach is suggested below.

Low R results in discovering larger attractors that generally cover a high percentage of the state space. For e.g. when only two subdivisions are used, an attractor corresponding to the whole state space may be discovered. The proposed method still safely identifies an attractor, but it only identifies the rough location of the attractor. A high R will identify more precisely the attractor, but may result in massive memory allocation. In the experimental evaluation, R is chosen such that the goal region has at least one cube of a uniform discretization.

The choice of H also has similar effects on the obtained description of the global dynamics. For small enough H, multi-valued map is closer to the identity map, resulting in the full state space being identified as the sole attractor. On the other hand, a large H may require more expensive computations, and the obtained outer approximation $\cF$ (digraph) may have more edges, making the topological ordering computationally expensive. In the experimental evaluation, H is selected with a short (wall clock) simulation time (ST) such that ST $\times$ R (the average total simulation time) is less than the time budget of the proposed experiment.


\section{Supportive Information for the Experimental Section}
\label{sec:appendix-experiments}

Table~\ref{tab:ground-truth} presents the parameters and results of the ground-truth evaluation for the different benchmarks. It reports three statistics: (a) The number of states in $X$ considered and the corresponding discretization of the state space ({\tt NS}). Unless indicated otherwise, $NS$ is computed by considering equal-spaced intervals of length $k$ along every dimension of $X$. (b) Time horizon provided to the controller to reach the goal condition or declare failure ({\tt H}) expressed in terms of simulation steps (=0.01seconds). And (c) Percentage of states in (a) from which the controller reached the goal condition ({\tt RoA ratio of $X$}).

\begin{table}[]
    \centering
    \vspace{-.25in}
    \begin{tabular}{|c||c|c|c|}
        \hline
        \textbf{Benchmark} & {\tt NS} & {\tt H} & {\tt RoA ratio of $X$} \\ \hline \hline
        Pendulum (LQR) & 3.1M = $1257 \times 2514$ & 500 & 38.63\%  \\ \hline
        Pendulum (Learned) & 3.1M = $1257 \times 2514$ & 500 & 18.56\% \\ \hline
        Acrobot (LQR) & $58.1M = 63 \times 63 \times 121 \times 121 $ & 10k & 100\% \\ \hline
        Acrobot (Hybrid) & $625k = 50^4$ & 5k & 100\% \\ \hline
        Ackermann (LQR) & 20.1M $= 401 \times 401 \times 126$ & 1k & 8.6\% \\ \hline
        Ackermann (Corke) & 20.1M $= 401 \times 401 \times 126$ & 1k & 70.65\% \\ \hline
        Ackermann (Learned) & $1M = 100 \times 100 \times 100$ & 1k & 100\% \\ \hline
    \end{tabular}
    \caption{\small Parameters and results of the ground truth RoA evaluation.}
    \vspace{-.25in}
    \label{tab:ground-truth}
\end{table} 

Table~\ref{tab:morse-experiments} presents the parameters used for executing the propagations of the dynamical system needed by the {\tt Morse Graph} approach. 

\begin{table}[]
    \centering
    \vspace{-.25in}
    \begin{tabular}{|c||c|c|}
        \hline
        \textbf{Benchmark} & {\tt NS} & {\tt H} \\ \hline \hline
        Pendulum (LQR) & $2^{16} = 65,536$ & 100 \\ \hline
        Pendulum (Learned) & $2^{16} = 65,536$ & 100 \\ \hline
        Acrobot (LQR) & $2^{20} = 1,048,576$ & 1100 \\ \hline
        Acrobot (Hybrid) & $2^{20} = 1,048,576$ & 5k \\ \hline
        Ackermann (LQR) & $2^{20} = 1,048,576$ & 500 \\ \hline
        Ackermann (Corke) & $2^{20} = 1,048,576$ & 500 \\ \hline
        Ackermann (Learned) & $2^{20} = 1,048,576$ & 500 \\ \hline
    \end{tabular}
    \caption{\small Parameters and results of the {\tt Morse Graph} RoA evaluation.}
    \vspace{-.25in}
    \label{tab:morse-experiments}
\end{table}




\end{document}

%% file: defs.tex
\newcommand{\cF}{{\mathcal F}}

\newcommand{\cM}{{\mathcal M}}

\newcommand{\cX}{{\mathcal X}}

\newcommand{\sM}{{\mathsf M}}

\newcommand{\sO}{{\mathsf O}}
\newcommand{\sP}{{\mathsf P}}

\newcommand{\sT}{{\mathsf T}}

\newcommand{\sCG}{\mathsf{CG}}
\newcommand{\smRoA}{\mathsf{mRoA}}

    \newfont{\mbf}{msbm10 scaled 1100}

\newcommand{\mvmap}{\rightrightarrows}

\newcommand{\sMG}{{\mathsf{ MG}}}

\newcommand{\sABlock}{{\mathsf{ ABlock}}}

%% file: correctm.tex
\def\corcommstyle{\bf\small\tt}

\def\corrl #1<<#2||#3>>{
\if\visiblecomments y
  \begin{quote} {\corcommstyle $<<$COMMENT$>>$ {\color{red}#1\marginpar{!!}}\\$<<$OLD$<<$} \end{quote}

{\color{red} 
 #2
 }

  \begin{quote} {\corcommstyle ==NEW== } \end{quote}
   \noindent\hrulefill
 
\vspace{-10pt} 
 
 \noindent\hrulefill
 
 \vspace{-10pt} 
 
 \noindent\dotfill
 
  #3
  
   \noindent\dotfill 

\vspace{-10pt} 
 
 \noindent\hrulefill
 
 \vspace{-10pt} 
 
 \noindent\hrulefill
  \begin{quote} {\corcommstyle $>>$END$>>$ } \end{quote}
 \else
  #3
 \fi
}

\long\def\longcorrl #1<<#2||#3>>{
\if\visiblecomments y
  \begin{quote} {\corcommstyle $<<$COMMENT$>>$ {\color{red}#1\marginpar{!!}}\\$<<$OLD$<<$} \end{quote}
 
 {\color{red}

  #2
  
  }
  
  \begin{quote} {\corcommstyle ==NEW== } \end{quote}
  
    \noindent\hrulefill
 
\vspace{-10pt} 
 
 \noindent\hrulefill
 
 \vspace{-10pt} 
 
 \noindent\dotfill
 
  #3
  
   \noindent\dotfill 

\vspace{-10pt} 
 
 \noindent\hrulefill
 
 \vspace{-10pt} 
 
 \noindent\hrulefill
  \begin{quote} {\corcommstyle $>>$END$>>$ } \end{quote}
 \else
  #3
 \fi
}

\def\mlabel #1
{
  \if\visiblecomments y
     \marginpar[\flushright \bf \footnotesize #1]{\bf \footnotesize #1}
  \fi
}

\def\flabel #1
{
  \if\visiblecomments y
       \hbox{\bf\footnotesize #1}
  \fi
}

\def\corrq #1<<#2>>{
\if\visiblecomments y
  \begin{quote} {\corcommstyle $<<$COMMENT$>>$ #1\marginpar{!!}\\$<<$BEG$<<$} \end{quote}
  \noindent\hrulefill
 
\vspace{-10pt} 
 
 \noindent\hrulefill
 
 \vspace{-10pt} 
 
 \noindent\dotfill
 
 {\color{red}
  
  #2
  
  }
   
  \noindent\dotfill 

\vspace{-10pt} 
 
 \noindent\hrulefill
 
 \vspace{-10pt} 
 
 \noindent\hrulefill 
  \begin{quote} {\corcommstyle $>>$END$>>$ } \end{quote}
 \else
  #2
 \fi
}

\long\def\longcorrq #1<<#2>>{
\if\visiblecomments y
  \begin{quote} {\corcommstyle $<<$COMMENT$>>$ #1\marginpar{!!}\\$<<$BEG$<<$} \end{quote}
  \noindent\hrulefill
 
\vspace{-10pt} 
 
 \noindent\hrulefill
 
 \vspace{-10pt} 
 
 \noindent\dotfill

  #2

  \noindent\dotfill 

\vspace{-10pt} 
 
 \noindent\hrulefill
 
 \vspace{-10pt} 
 
 \noindent\hrulefill 
  \begin{quote} {\corcommstyle $>>$END$>>$ } \end{quote}
 \else
  #2
 \fi
}

\def\corrc #1<<>>{
\if\visiblecomments y
  \begin{quote} {\corcommstyle $<<$COMMENT$>>$ \color{red} #1\marginpar{!!}} \end{quote}
\fi
}

\def\corre #1<<#2||#3>>{
\if\visiblecomments y
  #3\marginpar{\corcommstyle #1}
 \else
  #3
 \fi
}

\long\def\longcorre #1<<#2||#3>>{
\if\visiblecomments y
  #3\marginpar{\corcommstyle #1}
 \else
  #3
 \fi
}

\def\corrs #1<<#2||#3>>{
\if\visiblecomments y
  #3\marginpar{\corcommstyle #2 $\rightarrow$ #3\\ #1}
 \else
  #3
 \fi
}

\def\corro #1<<#2||#3>>{
#2}

\def\corrn #1<<#2||#3>>{
#3}

\long\def\longcorro #1<<#2||#3>>{
#2}

\long\def\longcorrn #1<<#2||#3>>{
#3}

\long\def\underconstruction #1<<<#2>>>{
\if\visiblecomments y
  \begin{quote} {\corcommstyle $<<$UNDER CONSTRUCTION - BEGIN$>>$ #1\marginpar{!!}} \end{quote}
  #2
  \begin{quote} {\corcommstyle $>>$UNDER CONSTRUCTION - END$>>$ } \end{quote}
 \else
 \fi
}

\def\showcomments{
  \let\visiblecomments y
}

\def\hidecomments{
  \let\visiblecomments n
}
